\pdfoutput=1

\documentclass[11pt]{article}

\usepackage{emnlp2021}

\usepackage{times}
\usepackage{latexsym}
\usepackage{enumitem}
\usepackage{comment}
\usepackage[T1]{fontenc}

\usepackage[utf8]{inputenc}
\usepackage{tabularx}

\usepackage{booktabs}
\usepackage{graphicx}
\usepackage{microtype}
\usepackage{amsmath}
\usepackage{mathtools}
\usepackage{amsfonts}

\DeclareMathOperator*{\argmin}{arg\,min}
\usepackage{adjustbox} 

\newcommand{\newpara}[1]{\noindent \textbf{#1} \hspace{0.5em}}

\usepackage{multirow}
\usepackage[makeroom]{cancel}
\usepackage{xcolor}
\newcommand{\ours}{\textsc{DePeN}}
\usepackage{hyperref}
\usepackage{cleveref}
\crefname{figure}{Figure}{Figure}
\crefname{table}{Table}{Table}
\crefname{section}{Section}{Section}
\crefname{appendix}{Appendix}{Appendix}
\title{\ours: Detect and Perturb to Neutralize Sensitive Attributes}

\title{
Detect and Perturb: Neutral Rewriting of Biased and Sensitive Text\\ via Gradient-based Decoding
}

\author{Zexue He, Bodhisattwa Prasad Majumder, Julian McAuley \\
  Department of Computer Science and Engineering, 
  UC San Diego  \\
  \texttt{\{zehe, bmajumde, jmcauley\}@eng.ucsd.edu} \\}

\begin{document}
\maketitle
\begin{abstract}
Written language carries 
explicit and implicit biases that can distract from meaningful signals. For example, letters of reference may describe male and female candidates differently, or their writing style may indirectly reveal demographic characteristics. At best, such biases distract 
from the meaningful content of the text; at worst they can lead to unfair outcomes.
We investigate the challenge of re-generating 
input sentences 
to
`neutralize'
sensitive attributes while 
maintaining the semantic meaning of the original text (e.g.~is the candidate qualified?). We propose a gradient-based rewriting framework, \textbf{De}tect and \textbf{Pe}rturb to \textbf{N}eutralize (\ours), that first detects sensitive components 
and masks them for regeneration, then perturbs the generation model at decoding time under a \textit{neutralizing} constraint that pushes the (predicted) distribution of sensitive attributes
towards a uniform distribution.
Our experiments in two different scenarios 
show
that \ours\ can regenerate fluent alternatives that are neutral in the sensitive attribute while maintaining the semantics 
of other attributes.
\end{abstract}

\section{Introduction}
Language data often 
carries implicit biases or contains sensitive information
that may 
have negative
consequences for human and machine understanding. For example, a person's choice of vocabulary can reveal their social identity (age, gender, or
political affiliation) \cite{nguyen2013old}; 
a few
examples are shown in \cref{tab:example1}. Such information 
can potentially bias machine predictions
as well as human judgment,
leading to unfair outcomes.

Hiding sensitive information 
in
textual data---including text that 
carries \emph{implicit} bias---is an essential task.
In this paper we
consider 
the setting of
graduate school admissions as a case-study, where 
fair evaluation of applicants should depend 
on academic performance or research potential, 
irrespective of nationality, gender, etc.
Text from reference letters is colored by many biases: letter writers may (possibly unintentionally) write about male and female candidates differently, or may 
use language that reflects
their (the writer's or the applicant's) cultural background.
Eliminating these attributes from the decision making process is challenging because (1) the sensitive information is 
often
implicit and 
confounded with other attributes,
and (2) a parallel corpus with 
\emph{unbiased} text is not available.

\begin{table}[t!]
\small
\begin{tabularx}{\linewidth}{X l}
\toprule
\bf Text    & \bf Attr. \\ \midrule
1. She \colorbox{pink}{gone} dance without da bands lol. & Race    \\
2. \colorbox{pink}{Hahaahhahaha} \colorbox{pink}{wwatching} rtl gemist holland, bigga is \colorbox{pink}{cryingg} it’s \colorbox{pink}{killinggg} me. &  Age \\
3. Tasted as \colorbox{pink}{amazing} as the first sip I took! \colorbox{pink}{Definitely} would recommend& Gender    \\
4. PERSON-B-1 is \colorbox{pink}{adorable} with pleasant and easy-going personality.    & Gender    \\ \bottomrule
\end{tabularx}
\caption{\small Examples of 
scenarios that reveal sensitive attributes (Attr.). 
Highlighted
words are 
markers of such sensitive information. Example 1 shows an excerpt of a tweet written by an African-American revealed by vocabulary usage (future tense of \textit{gone} $\rightarrow$ ``is going to'') \cite{blodgett2018twitter}. Example 2 is a tweet from a young person \cite{nguyen2013old}. Example 3 is a review by a female (from Yelp dataset \cite{reddy2016obfuscating}) while Example 4 describes a female applicant in a graduate admission reference letter (our data).}
\label{tab:example1}
\vspace{-1em}
\end{table}

Based on these motivations, we define our task as: given an input sentence associated with
both meaningful and sensitive attributes (e.g.~a discussion of a female student's research potential),
\textit{re-generate} 
the input in a way that
\textit{neutralizes} 
one or many \textit{sensitive attributes} with \textit{minimal edits}, i.e.,~so as to maintain
the fluency, coherency, and semantic meaning of the original sentence.


\begin{figure}[h] 
\centering 
    \includegraphics[width=0.95\linewidth]{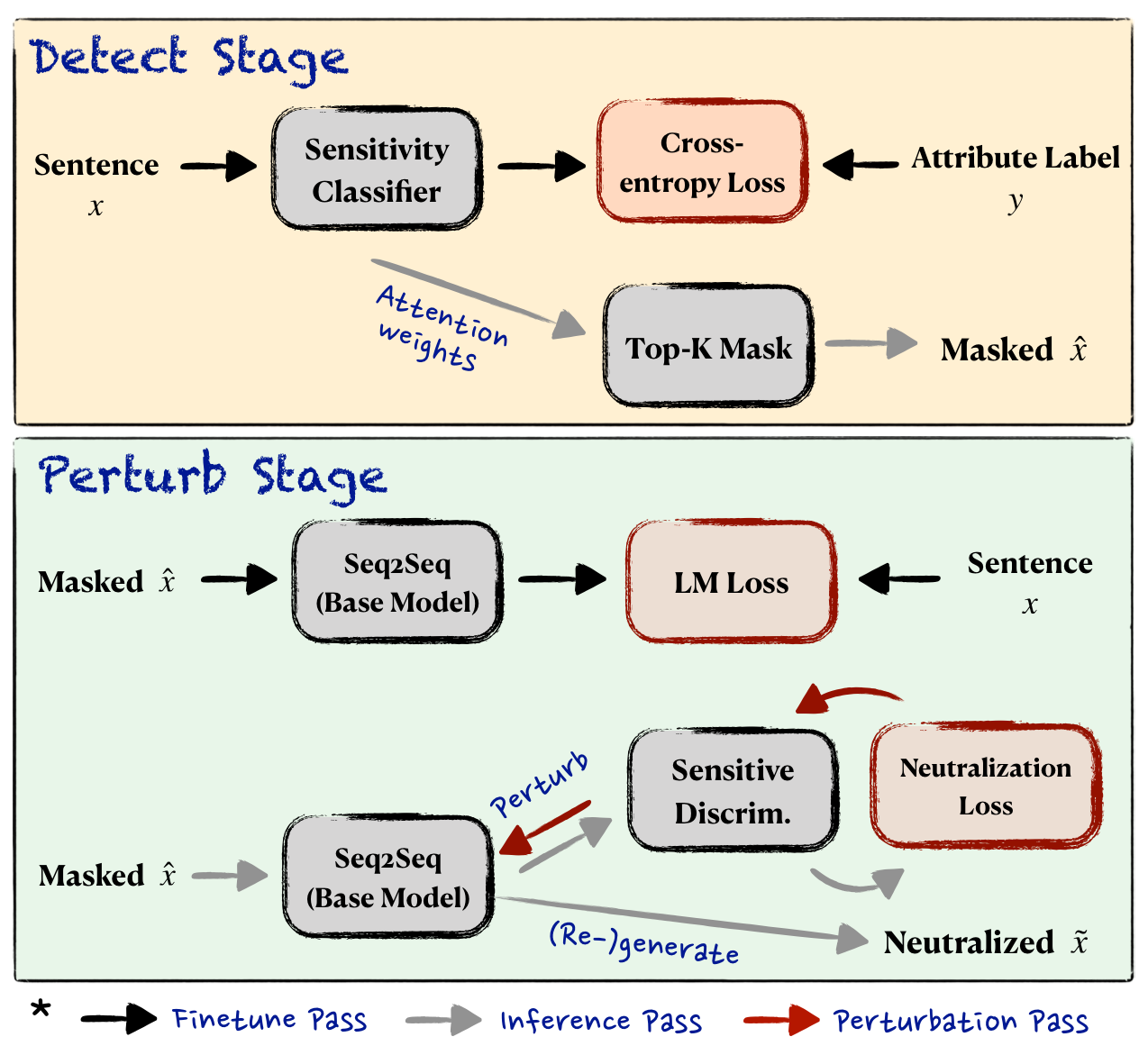}
  \caption{\small The dataflow of \ours. Details 
  of the Detect stage and Perturb stage are explained 
  in \cref{sec: methodology}. }
  \label{fig:framework}
  \vspace{-1em}
\end{figure}

To this end, we propose a gradient-based decoding framework for
text re-generation by neutralizing a sensitive attribute: \textbf{De}tect and \textbf{Pe}rturb to \textbf{N}eutralize (\ours).
We realize the framework in two steps (\cref{fig:framework}).
First we automatically detect the parts of the input sentence that 
reveal
the sensitive attribute, and mask them; 
while this can be as simple as a gendered pronoun (`he/she'), we find many cases where 
choices of adjectives or phrasing are associated with group identity.
Second, we regenerate a complete sentence from the unmasked part of the input so that the output no longer reveals 
the sensitive attribute. We do this by perturbing the final hidden states of a conditional language model that is finetuned to generate a complete sentence from masked tokens. Perturbation is done to modify the hidden states in a `neutral'
(i.e.,~so that the hidden state cannot predict the sensitive attribute)
direction 
while maintaining 
fluency and semantic meaning. We conduct two experiments to show that \ours~generalizes across scenarios. We first experiment with a Graduate Admissions Reference letter dataset where \ours~rewrites the sentences from a letter to neutralize attributes such as gender or nationality. 
So that we can release a
reproducible benchmark,
we also experiment with \emph{Goodreads} review data \cite{wan2018item};
here we 
treat
genres as a sensitive attribute (i.e.,~maintain the essence of a review without revealing the genre). 

\section{\ours}
\label{sec: methodology}
As shown in \cref{fig:framework}, our neutralizing approach \ours\footnote{\url{https://github.com/ZexueHe/DEPEN}.}~has two stages: Detect and Perturb.  

\subsection{Detect: mask the sensitive parts}
First we
detect parts of the original input sentence $x$ that are
predictors of the target sensitive attribute $\mathcal{A}$.
Suppose we have a corpus containing $N$ documents and their associated label $y$ for $\mathcal{A}$;
we train a classifier $f_\theta$ to $
\underset{\theta}{\text{minimize }}\frac{1}{M}\sum_{i=1}^{N}\sum_{j=1}^{|X^i|}\mathcal{L}(f(x^i_j; \theta), y^i)
$, where $X^i$ is the $i$-th document and $x^i_j$ is the $j$-th sentence, $M$ is  the number of sentences,
and $\mathcal{L}$ is the cross-entropy loss for classifying sensitive attributes.

Following \citet{jain2020learning}, we take 
self-attention scores of all input tokens w.r.t.~the \texttt{[CLS]} token \cite{devlin2018bert}  from the final hidden layers and normalize them to measure how salient each token is for predicting $\mathcal{A}$. 
We use BERT as the attribute classifier $f$.

Next, we mask the top-$k\%$ ($k$ is a hyperparameter) salient tokens to obtain the intermediate output as $\hat{x}^i_j$ that does not contain any significant predictor of $\mathcal{A}$ according to $f$.

\subsection{Perturb to Neutralize}
To regenerate a neutral version $\Tilde{x}$ of the original input sentence $x$ we need a generative model that can reconstruct a sentence from the unmasked tokens.
For this we train a sequence-to-sequence (Seq2Seq) model 
that takes $\hat{x}^i_j$ as input and $x^i_j$ as output.
We finetune a BART model as our base Seq2Seq model $g$. Ideally, we want $g$ to regenerate a version that remains neutral to the attribute $\mathcal{A}$. But since we do not have attribute-neutral 
ground-truth,
we cannot guarantee that inference from $g$ will hold attribute neutrality. Hence, we guide $g$ using a gradient-based inference method so that the regenerated output remains attribute-neutral. We are inspired by PPLM \cite{dathathri2019plug} that introduced gradient-based inference from transformer-based language models. Similar inference-time perturbation approaches also have been proposed for applications such as clarification question generation \cite{DBLP:conf/naacl/MajumderRGM21} and dialog generation \cite{DBLP:conf/acl/MajumderBMJ20}.

PPLM primarily performs gradient-based decoding that encourages the generation to maintain fluency according to the base autoregressive generative model while honoring a discriminative constraint, such as maintaining a particular attribute. In our work, we modify PPLM to accommodate a new decoding constraint for achieving neutrality. We also adapt a Seq2Seq transformer model as a base model to perform autoregressive inference using PPLM-style gradient decoding. 

\paragraph{Generate with Neutralizing Constraints} 
Contrary to PPLM, which boosts the log-likelihood (LL) of a certain attribute,
our case requires the generation 
is \emph{neutral} toward an attribute (e.g.~the text should be
neither `female' nor `male'). Since we do not have explicit labels for 
neutrality,
we 
modify our decoding constraint in the following. 

Suppose there are $|\mathcal{C}|$ categories for  $\mathcal{A}$ and we want to re-generate a sentence $\Tilde{x}_j^i$ which minimizes the KL-divergence between a uniform distribution over $\mathcal{C}$  and the discriminative distribution of the sensitive attribute $\mathcal{A}$. We define it as our neutralization constraint $\mathcal{L}_{\text{ntrl}}$ 
\begin{equation*}
    \begin{aligned}
    &\underset{\Tilde{x}_j^i}{\argmin}\ D_{KL}\left(U(\mathcal{C})\ ||\ p(y^i|\Tilde{x}_j^i)) \right) \\
    = &\ \underset{\Tilde{x}_j^i}{\argmin}\ H\left(U(\mathcal{C}), p(y^i|\Tilde{x}_j^i)\right) - \bcancel{H\left(U(\mathcal{C})\right)}\\
    =&\ \underset{\Tilde{x}_j^i}{\argmin} \underbrace{-\sum_{a \in \mathcal{C}}\frac{1}{|\mathcal{C}|}\log p(y^i =a |\Tilde{x}_j^i)}_{ \mathcal{L}_{\text{ntrl}}}
    \end{aligned}
\end{equation*}
where $H(\cdot)$ is the entropy and $U(\cdot)$ denotes the uniform distribution.


Since ground truth is not available,
we resort to an unsupervised decoding technique using the left-to-right decoder from the Seq2Seq model.
During inference, we keep the encoder of the base model $g$ fixed while perturbing the hidden states of the decoder.
A gradient w.r.t.~the neutralization loss $\mathcal{L}_{\text{ntrl}}$ 
shifts the hidden state representations toward neutrality during backpropagation. To realize the effect of backward gradient updates, we accumulate gradients for multiple passes and then update the hidden representations. Once we update decoder hidden states, a forward pass is made to 
maintain the fluency of the base language model. Backward and forward pass alternate 
until we see the desired neutralization effect 
in the generated text. 
\section{Experiments}

\subsection{Datasets}
\paragraph{Reference letters} a real-world dataset of students considered for admission to a graduate program of a large US university,\footnote{Our investigation is IRB-approved. Details are anonymized even in our private version.} containing applicant profiles including reference letters, binary gender information, nationality, and a binary admission decisions. We 
consider
18,865 applicants with 29,170 reference letters,
among which 
22,201 letters are used for training classifiers and 6,969 for testing or rewriting. We conduct two experiments with gender and nationality (processed to be 4 dominant classes) as sensitive attributes separately, and use admission decisions as the outcome for further evaluating 
whether the `signal' is preserved.
\paragraph{GoodReads} a book review
dataset \cite{wan2018item} containing user reviews, star ratings, and genres.
We randomly sample 3000 reviews each from the \emph{Children's} 
and 
\emph{Mystery}
genres.
We use 5000 reviews 
for training and the rest for testing.
We define the binary genre as the sensitive attribute, and
quantize ratings to three levels (positive, negative, neutral)
as the outcome.

\begin{table*}[t!]
\small
\resizebox{\textwidth}{!}{%
\begin{tabular}{cccccc|ccccc|ccccc}
\toprule
\multirow{3}{*}{\bf Model}                                            & \multicolumn{10}{c}{\bf CS Admission Dataset}                                                                                                                                           & \multicolumn{5}{c}{\bf GoodReads Dataset}                                                    \\ \cmidrule(l){2-11}\cmidrule(l){12-16} 
                                                                  & \multicolumn{5}{c}{Gender (binary)}                                                      & \multicolumn{5}{c}{Nationality (4 classes)}                                              & \multicolumn{5}{c}{Genre (binary)}                                                       \\ \cmidrule(l){2-6}\cmidrule(l){7-11} \cmidrule(l){12-16} 
                                                                  & Acc.            & Conf.           & PLL              & BLEU4           & Out.            & Acc.            & Conf.           & PLL              & BLEU4           & Out.            & Acc.            & Conf.           & PLL              & BLEU4           & Out.            \\ \midrule
\begin{tabular}[c]{@{}c@{}}Original\end{tabular}          & 0.9247          & 0.9002          & -4.8134          & 1.0000          & 0.6321          & 0.7487          & 0.6660          & -4.6511          & 1.000           & 0.6741          & 0.7557          & 0.7165          & -4.3154          & 1.0000          & 0.6551          \\ \cmidrule(l){1-16}
\begin{tabular}[c]{@{}c@{}}RB\end{tabular} & 0.7397          & 0.6614          & -5.0973          & 0.8761 & 0.6333 & 0.7470          & 0.6665          & -4.8624          & 0.9974 &  0.6353    & 0.7297          & 0.6938          & -4.4106          & 0.9699 & 0.6543 \\
\begin{tabular}[c]{@{}c@{}}WD\end{tabular}      & 0.7125          & 0.6940          & -5.0520          & 0.3781          & 0.6101          & 0.6303          & 0.5568          & -4.6771          & 0.5251          & 0.6105          & 0.6565          & 0.6885          & -4.5162          & 0.2571          & 0.5905          \\
\begin{tabular}[c]{@{}c@{}}ADV\end{tabular}   & 0.9197                &   0.8970              &  -5.9049                &  0.3979               &    0.5818             & 0.7091          & 0.6302          & -5.8053          & 0.3838          & 0.5551          & 0.7364          & 0.7149          & -4.7013          & 0.2917          & 0.5978          \\
\begin{tabular}[c]{@{}c@{}}PATR\end{tabular}   & 0.8797                &   0.8528              &  -5.0034                &  0.5278               &    0.6071             & 0.7148          & 0.629          & -4.7511          & 0.5336          & 0.6154          & 0.7451          & 0.7077          & -4.4254          & 0.3637          & 0.5979          \\
\cmidrule(l){1-16}
\textsc{DeN}                                                               & 0.7546          & 0.7375          &  -4.8695    & 0.4449          & 0.6171          & 0.6416          & 0.5694          &  -4.6696    & 0.5261          & 0.5818          & 0.6815          & 0.6534          & -3.9767 & 0.3749          & 0.6015          \\
\textsc{PeN}                                                               & 0.5002 & 0.4617 & -5.1048          & 0.0825          & 0.5871          & 0.2486 & 0.2489 & -5.0260          & 0.0652          & 0.6245          & 0.5362 & 0.5219 & -4.9699          & 0.2471          & 0.5728          \\
\ours                                                             & 0.5157    &  0.4935    & -4.8464 & 0.6356    &  0.6298    & 0.5242    & 0.4564    & -4.6623 & 0.5357    & 0.6491 & 0.5915    & 0.5665    & -4.3798    & 0.3747    & 0.6168    \\ \bottomrule
\end{tabular}%
}
\caption{\small Results on 
Reference Letters and GoodReads data
(see \cref{sec:analysis}).
} 
\label{tab:results}
\end{table*}

\subsection{Evaluation Metrics}
\textbf{Bias:}
We use the accuracy (Acc.) and confidence (Conf.) 
of a sensitive classifier to evaluate bias. 
\textbf{Fluency:}
We use the Pseudo Log-Likelihood (PLL) of \citet{salazar2019masked} to measure the fluency of our generated model. 
\textbf{Coherence:}
We use the BLEU4 score of the generated sentence w.r.t.~its input and accuracy of an outcome (Out.) classifier to measure how much content is maintained. 

\subsection{Baseline Models}
We evaluate four 
debiasing 
approaches
(all of which generate without parallel ground truth)
and two variants of \ours\ as baselines:
\begin{itemize}[leftmargin=*,itemsep= 0.35 pt]
    \item Rule-based (RB): replace words with rules (e.g.~\emph{he/she} $\rightarrow$ \emph{they}, see \cref{sec:appendix rule-based}).
    \item Weighed Decoding (WD): a 
    decoding method \cite{ghazvininejad2017hafez} by reducing the generation probability of detected sensitive tokens to a hyperparameter $\alpha$ (we set $\alpha=0.2$).
    \item Adversarial Training (ADV): a Seq2Seq autoencoder with a gradient reversal layer \cite{ganin2015unsupervised} that propagates gradients of the sensitive discriminator to the encoder.
    \item Privacy-Aware Text Rewriting (PATR): we reimplement the adversarial back-translation rewriting model of \citet{xu2019privacy}. 
    \item \textsc{DeN}: \ours\ w/o Perturb, generates $\Tilde{x}$ from $\hat{x}$ with the finetuned base model $g$.
    \item \textsc{PeN}: \ours\ w/o Detect, generates $\Tilde{x}$ from $x$ by neutrally perturbing a normal Seq2Seq. 
\end{itemize}
\subsection{Results and Analysis}
\label{sec:analysis}
Results are shown in \cref{tab:results}. 
For debiasing metrics, \ours\ 
leads to a decrease (as desired) in
Acc. and Conf. 
to around 0.5 for all experiments.
We note
that \textsc{PeN} generates sentences with a normal BART designed for common Seq2Seq tasks like summarization or translation, so 
in spite of a somewhat better accuracy drop,
regenerated sentences 
differ vastly
from inputs,
which can be seen from low BLEU4 scores (0.0825 for gender and 0.06 for nationality). 
WD also 
lowers bias, but 
it can abruptly interrupt the generation by reducing the probabilities of certain (sensitive) tokens affecting the overall language model fluency. 



We also report the accuracy of predicting outcome variables (Out.), i.e.,~admission decisions or review sentiment (which are \emph{not} used for training).

For fluency 
\textsc{DeN} 
has the highest (i.e.,~best) PLL but 
fails to debias (high Acc.~and Conf.). \textsc{DePeN} maintains high fluency while also debiasing.

RB has the highest coherence, 
though 
we find
that
regenerated sentences
are extremely similar to the input (with many biased terms persisting) due to simple replacement rules.
RB has extremely high BLEU4 scores (0.9974 for nationality and 0.9699 for GoodReads). PATR also demonstrates its effectiveness on language quality (fluency and coherence) due to the paraphrasing capability of back-translation, however it fails to debias well as it still shows high Acc.~and Conf.~in bias classification (more in \cref{sec:appendix 1}).

\ours~beats the baselines by achieving a 
balance across bias mitigation, fluency, and coherency, and fidelity w.r.t.~the predicted outcome. 
Manual inspection revealed that automatic metrics are suggestive of how humans perceive neutrality.

\subsection{Case Study}
We provide an example in \cref{tab:case study}, in which a referrer comments on the mock classes of a student. More examples and findings are shown in 
\cref{sec:appendix 1}.
Besides the obvious gendered indicators \emph{Her/girl}, the words \emph{lovely} and \emph{popular} are also considered as 
gender-predictive. For RB, such 
adjectives 
strain the ability of humans
to design perfect rules, not only because it is hard to enumerate all such words but 
also due to their context-dependence
(e.g.~`elegant' may carry different bias if it describes a student versus a student's theorem).
Simple 
replacement (e.g.~\emph{their}) also yields ungrammatical sentences. For WD and \textsc{DeN}, without a neutralization constraint, they 
select
candidates 
that satisfy the
language model, 
but
may choose (e.g.)~\emph{man},
leading to no reduction in attribute sensitivity, and (e.g.)~\emph{active} which changes the semantic meaning.
As a black-box rewriting method with strong reconstruction signals, it's harder to control ADV to meet all expectations simultaneously. PATR also fails to debias. However, \ours\ can edit the sensitive parts 
while maintaining
fluency and semantic meaning. 
\begin{table}[t!]
\scriptsize
\resizebox{\linewidth}{!}{
\begin{tabularx}{\linewidth}{c X }
\toprule
Model &  Re-generated \\ \midrule
 Original             & \colorbox{pink}{Her} course really attracted others, that made this \colorbox{pink}{lovely} \colorbox{pink}{girl} really \colorbox{pink}{popular} in classroom.
 \\ \cmidrule(l){1-2}
 RB                   & \emph{Their} course really attracted others, that made this \emph{lovely person} really \emph{popular} in classroom.
                                                                                                       \\
WD                   & \emph{The} course instantly attracted others, that made this \emph{young man} really \emph{active} in classroom.                                                                                \\
ADV                 & \emph{Her} course really attracted others, that made this \emph{excited girl} really \emph{popular} in classroom.                                                                                                        \\
PATR                 & \emph{Her} course really attracted others, which made this \emph{lovely girl} really \emph{popular} in class.                                                                                                   \\
\textsc{DeN}                   & \emph{The} course instantly attracted others, that made this \emph{young man} really \emph{active} in classroom                                                                                   \\
\textsc{PeN}                  &  A class almost one third time I got on the topic, but it's true for the classroom at home.                                                   \\
\ours                & \emph{This} course instantly attracted others, that made this \emph{young student} really \emph{shine} in classroom.                                                                            \\ \bottomrule
\end{tabularx}
}
\caption{\small Re-generated examples. 
We show detected sensitive words in red, and edited words in italics.}
\vspace{-1em}
\label{tab:case study}
\end{table}

\section{Related Work}
\newpara{Debiasing Language Generation} There are three main
streams to debias NLG tasks:  counterfactual data augmentation 
\cite{lu2020gender,chen2018learning};  training-time methods \cite{huang2019reducing, liu2019does, liu2020mitigating, kaneko2021debiasing, pryzant2020automatically}; and inference-time methods.
\citet{saunders2020reducing} mitigate gender bias in machine translation via transfer learning 
using handcrafted
gender-balanced datasets. \citet{sheng2020towards} generate with well-formulated bias triggers based on \cite{wallace2019universal} to equalize biases between demographics. \citet{dathathri2019plug} propose a gradient-based method for controllable generation and show its 
efficacy
in toxicity reduction. However, all these methods require 
explicit labels or parallel data regarding the desired attribute.

\newpara{Re-writing} 
Here
specific parts of the original text are revised to be more aligned with a target attribute \cite{thompson2013autism}. 
Representative approaches use
an encoder-decoder setup with a discriminator (e,g.~style) \cite{romanov2018adversarial, dai2019style,john2018disentangled, Aho:72, DBLP:conf/acl/MajumderBMJ20, DBLP:conf/naacl/MajumderRGM21}, 
backtranslation \cite{lample2018multiple,prabhumoye2018style, xu2019privacy},  pretraining \cite{duan2020acl,zhou2021emnlp}, or use retrieval framework \cite{sudhakar2019transforming}. 
A few approaches adapt these techniques for debiasing.
\citet{zmigrod2019counterfactual} mitigate gender bias by converting between masculine- and feminine-inflected sentences with data augmentation;
\citet{ma2020powertransformer} jointly train a reconstruction and an out-of-domain paraphrasing task to correct bias, which requires a parallel corpus with
attribute-sensitive (e.g.~gender) verbs assigned and masked. 
In contrast,
we aim to rewrite neutrally without human 
guidance.

\section{Conclusion}
In this work, we propose a gradient-based rewriting framework, \ours, to neutralize a text that carries sensitive information (e.g., gender) by detecting the sensitive-predictable parts and perturbing the regeneration via a neutralization constraint. The constraint will shift the re-generated sentences to be uniform distributed for the sensitive attribute (e.g., neither male nor female) with minimal editing to maintain the semantic content.

\section*{Acknowledgments}
We thank Taylor Berg-Kirkpatrick, Jianmo Ni, Yuheng Zhi, and anonymous reviewers for their valuable suggestions to this work. Our Reference Letter dataset is built from anonymized admission data of UC San Diego's CSE department, and its use is supported by our IRB. 
BPM is partly supported by a Qualcomm Innovation Fellowship and NSF Award \#1750063.

\section*{Ethical considerations}

While a debiasing system is intended to mitigate fairness issues in natural language, such a system could certainly have unwanted side effects. Most critically, removing bias may to some extent eliminate meaningful signal from the data, or subtly alter the intended meaning of a sentence. A malicious user could adversarially maximize the neutralization constraint which would result in enhancing the bias in the input sentence.
A system like ours should likely not be used as a `black box,' but would best be used in a setting where its outputs can be `audited' to ensure that semantic meaning is preserved, e.g.~by a letter writer trying to improve their own writing or by a neutral third party.

\bibliography{anthology,custom}
\bibliographystyle{acl_natbib}
\clearpage

\appendix

\section{Appendix}
\subsection{Details about Baselines}
\label{sec:appendix rule-based}
\paragraph{Rule-based Model} Detailed rules are described in \cref{tab:replacement rule}. For gender, we follow the handout\footnote{\url{https://writingcenter.unc.edu/tips-and-tools/gender-inclusive-language/}} for mitigation. For nationality, though we have masked the sensitive information with Named Entity Recognition (NER), 
there are a few 
cases where NER fails, such as ``Chinese Mathematical Olympiad''; 
to handle this we delete a list of country/city/nationality names. Since we 
can't precisely
formulate the special patterns 
corresponding to
applicants from different nationalities,  we count 
unique bi-grams in the top-100 bi-gram list of each category as additional rules. For GoodReads, we use the listed featured words for mystery and children's books,\footnote{\url{https://www.vocabulary.com/lists/}} and handcraft their replacements. 

\paragraph{Privacy-Aware Text Rewriting (PATR)}
We re-implement \citet{xu2019privacy}'s adversarial rewriting model with Huggingface pretrained translators. We first translate English input to French\footnote{\url{https://huggingface.co/Helsinki-NLP/opus-mt-en-fr}} mediated results and translate it back to English\footnote{\url{https://huggingface.co/Helsinki-NLP/opus-mt-fr-en}}.


\subsection{Case Studies}
\label{sec:appendix 1}
\paragraph{Case Study 1 (Gender)}
In \cref{tab:adorable}, besides the pronoun \emph{her}, \emph{adorable} is also 
a strong
predictor of 
female gender 
(the
word \emph{`ributes'} is a typo 
by the referrer). 
Whether the adjectives are gendered depends on context
(e.g., ``beautiful work'' may not predict gender but ``beautiful person'' does).
This is a difficult case for 
RB 
and 
WD
to distinguish or to select the best replacements. 
ADV replaces the gendered but positive word \emph{adorable} with a neutral but less positive word \emph{third}. This 
reveals that while
ADV 
substitutes a less biased word, it
lacks 
the
ability to maintain the high-level semantic meaning. PATR shows its advantage of paraphrasing due to the back-translation, however, it fails to identify biased words and debias them.   

\textsc{DeN} and \ours\ successfully neutralize \emph{adorable} $\rightarrow$ \emph{commendable} or \emph{praiseworthy} which express not only the same semantic meaning but also the same high-level sentiment. 
Noting
that we don't have any sentiment guidance or constraint, this advantage is achieved by grasping the core content and inferring the underlying attitude. 
\textsc{DeN} and \ours\ can correct the 
typo
\emph{ributes} with a plausible replacement (\emph{work}).

Another interesting phenomenon is when \ours\ accidentally generates a gendered word (\emph{Her}), 
it compensates by correcting this to a proper noun (essentially an `invented name' 
\emph{Her $\rightarrow$ Heragur});
the new word still plays the same grammatical role in the sentence (e.g., \emph{Her} and \emph{Heragur's} are possessive pronouns with the same POS tag). This could perhaps be further improved by preventing the decoder from generating proper nouns at all, or otherwise by combining our decoding strategy with additional rules.

\paragraph{Case Study 2 (Gender)} As shown in \cref{tab:basketball}, \emph{He}, \emph{lover} and \emph{basketball} are 
predictive of (male) gender.
Although WD, ADV and \textsc{DeN} 
find a close
replacement (\emph{sports} for \emph{basketball}), the sentences still 
predict the male gender (they fail to correct the pronoun \emph{he}). While it replaces \emph{lover} with a more neutral word (\emph{enthusiast}), PART still generates \emph{he} and \emph{basketball}.
\textsc{PeN} 
again
rewrites the sentence in a way that differs drastically from the input.
\ours\ neutralizes the highlighted parts 
with
suitable replacements. 
\begin{table*}[t!]
\small
\resizebox{\textwidth}{!}{
\begin{tabularx}{\textwidth}{c X }
\toprule
\multicolumn{1}{l}{Sensitive Attr.} & \multicolumn{1}{c}{Rules}                                                                                                  \\ \midrule
\multirow{3}{*}{Gender}                 & \textbf{Replace} he/she $\rightarrow$ they,  his/him/her/hers $\rightarrow$ them/their, boy/girl $\rightarrow$ person                                               \\
                                        & \textbf{Delete} Mr., Ms., Miss, Mrs.                                                                                                   \\ 
                                        & \textbf{Replace} chairman/chairwoman$\rightarrow$ chair, actor/actress $\rightarrow$ actor,  freshman $\rightarrow$ first-year student ... \\\cmidrule(l){1-2}
\multirow{5}{*}{Nationality}            & \textbf{Delete} country/city/nationality names, e.g., China/Chinese, America/American, India/Indian, Taiwan ...      \\
                                        & Category 1:   \textbf{Replace} intellectual curiosity$\rightarrow${}ability                                                                    \\
                                        & Category 2: \textbf{Replace} solid foundation $\rightarrow$ understanding                                                                     \\
                                        & Category 3: \textbf{Replace} financial/finance aid/support/situation $\rightarrow$ support/situation                                          \\
                                        & Category 4: \textbf{Replace} senior project $\rightarrow$ project                                                                             \\\cmidrule(l){1-2}
\multirow{2}{*}{Genre}                  & \textbf{Replace} children/child/kid/boy/girl/daughter/son $\rightarrow$ reader,  picture/children/fairy book/story $\rightarrow$ book/story              \\
                                        & \textbf{Delete} murder, mystery, crime, suspect, suspense, victim, killer, investigation ...                                            \\ \bottomrule
\end{tabularx}%
}
\caption{Detailed replacement rules used in our rule-based baseline.}
\label{tab:replacement rule}
\end{table*}

\paragraph{Case Study (Nationality)}
\cref{tab:extracurricular} shows a sentence in a reference letter written for a US student. 
We find that `extracurricular' activities (both the word itself and the topic in general) tend to appear more in
letters for US (and to some extent Indian) students compared to (e.g.)~Chinese students; as such the word is detected as a predictor of nationality.
%
From \cref{tab:extracurricular}, although RB 
eliminates the indicator \emph{extracurricular}, it causes 
ambiguity by simply deleting it.
\ours\ replaces the indicator \emph{extracurricular} with \emph{social/cultural} which is not only semantically similar but also less predictive of nationality (note that the pronoun `her' is not removed from this sentence as it is not a sensitive attribute in this experiment).

\paragraph{Case Study (GoodReads)}
\cref{tab:chidren} shows a sentence 
from a
review of a children's book, where 
models rewrite to hide 
genre information while maintaining 
content (especially the review sentiment).
This example gives another 
illustration about why 
rule-based (RB) methods
fail: 
\emph{children} in 
this context
does not refer to the genre
but describes a specific character.
Distinguishing such differences 
would demand a more nuanced rule-based model, requiring significant handcrafting.
\ours\ can overcome this problem by doing inference automatically.

\begin{table*}[t!]
\small
\resizebox{\textwidth}{!}{
\begin{tabularx}{\textwidth}{c X }
\toprule
Model &  \multicolumn{1}{c}{Re-generated} \\ \midrule
 Original             & \colorbox{pink}{Her} desire for perfection, confidence levels, humility and excellent personal at \colorbox{pink}{ributes} are \colorbox{pink}{adorable}.          \\ \cmidrule(l){1-2}
 RB                   & \emph{Their} desire for perfection, confidence levels, humility and excellent personal at \emph{ributes} are \emph{adorable}.                                                                                                       \\
WD                   & \emph{Her} desire for perfection, confidence levels, humility and excellent personal at \emph{ributes} are \emph{adorable}.                                                                                \\
ADV                 & \emph{Her} desire for perfection, confidence levels, humility and excellent personal at \emph{kuutes} are \emph{third}.                                                                                                       \\
PATR                 & \emph{His} desire for perfection, level of confidence, humility and excellent personal \emph{attributes} are \emph{adorable}.                                                                                                       \\
\textsc{DeN}                   & \emph{Her} desire for perfection, confidence levels, humility and excellent personal at \emph{work} are \emph{commendable}.                                                                                   \\
\textsc{PeN}                  & Forges says, desire, humility, humility and desire to be honest with perfect personal.                                                    \\
\ours                &  \emph{Heragur's} desire for perfection, for perfection, confidence levels, humility and excellent personal at  \emph{work} are  \emph{praiseworthy}.                                                                             \\ \bottomrule
\end{tabularx}
}
\caption{Re-generated examples with gender as the sensitive attribute}
\label{tab:adorable}
\end{table*}

\begin{table*}[t!]
\small
\resizebox{\textwidth}{!}{
\begin{tabularx}{\textwidth}{c X }
\toprule
Model &  \multicolumn{1}{c}{Re-generated} \\ \midrule
 Original             & Meanwhile \colorbox{pink}{he} is not a keen \colorbox{pink}{lover} of \colorbox{pink}{basketball}, but also plays it with skills.          \\ \cmidrule(l){1-2}
 RB                   & Meanwhile \emph{they} is not a keen \emph{lover} of \emph{basketball}, but also plays it with skills.                                                                                                       \\
WD                   & Meanwhile \emph{he} is not a keen \emph{learner} of \emph{sports}, but also plays it with skills.                                                                                      \\
ADV                 & Meanwhile \emph{he} is not a keen \emph{learner} of \emph{sports}, but also plays it with skills.                                                                                                             \\ 
PATR                 &  Meanwhile \emph{he} is not a \emph{basketball enthusiast}, but also plays with skills.                                                                                                           \\
\textsc{DeN}                   & Meanwhile \emph{he} is not a keen \emph{lover} of \emph{sports}, but also plays it with skills.                                              \\
\textsc{PeN}                  & The Duchess of Amida Costa Rica plays the World No. 3-rank seven in a row.                                                      \\
\ours                &  Meanwhile: \emph{PERSON-I-2189} is not a keen \emph{learner} of \emph{sports}, but also plays it with skills.                                                                             \\ \bottomrule
\end{tabularx}
}
\caption{Re-generated examples with gender as the sensitive attribute.}
\label{tab:basketball}
\end{table*}

\begin{table*}[t!]
\small
\resizebox{\textwidth}{!}{
\begin{tabularx}{\textwidth}{c X }
\toprule
Model &  \multicolumn{1}{c}{Re-generated} \\ \midrule
 Original             & Apart from her \colorbox{pink}{studies} she has also taken keen interest in \colorbox{pink}{extracurricular} activities          \\ \cmidrule(l){1-2}
 RB                   & Apart from her \emph{studies} she has also taken keen interest in   activities .                                                                                                        \\
WD                   & Apart from her \emph{classes} she has also taken keen interest in \emph{co-curricular} activities.                                                                                     \\
ADV                 & Apart from her \emph{studies} she has also taken keen interest in \emph{extracurricular} activities.                                                                                            \\
PATR                 & Apart from her \emph{studies}, she also interested herself in \emph{extracurricular} activities.                                                                                            \\
\textsc{DeN}                   & Apart from her \emph{coursework} she has also taken keen interest in \emph{extra-curricular} activities.                                          \\
\textsc{PeN}                  & Samantha Smith is a great-and-groom at home to an amazing six-week-old.                                                                   \\
\ours                &  Apart from her \emph{coursework} she has also taken keen interest in \emph{social/cultural} activities                                                                             \\ \bottomrule
\end{tabularx}
}
\caption{Re-generated examples with nationality as the sensitive attribute.}
\label{tab:extracurricular}
\end{table*}

\begin{table*}[t!]
\small
\resizebox{\textwidth}{!}{
\begin{tabularx}{\textwidth}{c X }
\toprule
Model &  \multicolumn{1}{c}{Re-generated} \\ \midrule
 Original             & I didn't really get this one, although I liked the \colorbox{pink}{example} of 
 \colorbox{pink}{children} dealing with a new sibling.         \\ \cmidrule(l){1-2}
 RB                   & I didn't really get this one, although I liked the \emph{example} of \emph{readers} dealing with a new sibling.                                                                                                      \\
WD                   & I didn't really like this one, although I liked the \emph{story} about \emph{kids} dealing with a new sibling.                                                                                     \\
ADV                 & I didn't really get this one, although I liked the \emph{example} of \emph{children} dealing with a new sibling.                                                                                               \\
PATR                 &  I didn't really get this one, though I liked the \emph{example} of \emph{kids} dealing with a new \emph{brother and sister.}                                                                                             \\
\textsc{DeN}                   & I didn't really like this one, although I liked the \emph{idea} of \emph{siblings} dealing with a new sibling.             \\
\textsc{PeN}                  & Young Wolf stories that deal with siblings siblings.                                                    \\
\ours                &  I don't really like this one, although I liked the \emph{story} of \emph{characters} dealing with a new sibling.                                                                             \\ \bottomrule
\end{tabularx}
}
\caption{Re-generated examples with genre as the  sensitive attribute.}
\label{tab:chidren}
\end{table*}
\subsection{Data Preprocessing}
For the Reference Letter dataset, we first exclude invalid reference letters if the letter (1) is too short (less than 2 sentences), or (2) contains too many named entities (more than 90\%, presumably due to OCR problems), or (3) is not written in English.
For GoodReads dataset, we sample 3000 samples each from Childern's and Mystery's genre. 

\subsection{Details of Model}
\subsubsection{Number of Parameters}
In all experiments, we use BERT in the Detect stage, which has 110M parameters; we use BART as our base Seq2Seq model in the Perturb stage, which has 117M parameters. All classifiers are finetuned BERT.  

\subsubsection{Hyperparamters}
We use 64 as the batch size for finetuning all BERT classifiers and use 8 as the batch size for the BART  Seq2Seq model for finetuning or generation. We use AdamW\footnote{\url{https://pytorch.org/docs/master/generated/torch.optim.AdamW.html}} as the optimizer with initial learning rate of 1e-4. The whole pipeline is implemented with PyTorch\footnote{\url{https://pytorch.org/}}, and all transformers are implemented based on the libraries of Hugging Face \footnote{\url{https://huggingface.co/}}. 

In our Petrub stage,   we tried several $k$ ($k=10, 20, 30$) during our implementation and we found that our results are not sensitive to the choice of $k$.

\subsection{Details of Datasets}
We download the GoodReads book review dataset by genre from the official website\footnote{\url{https://sites.google.com/eng.ucsd.edu/ucsdbookgraph/home\#h.p_VCP_qovwtnn1}}.  
\subsection{Details of Evaluation Metrics}
\href{https://www.nltk.org/api/nltk.translate.html}{ \texttt{nltk.translate.bleu\_score.corpus\_\\bleu}} from \texttt{nltk} package is used to calculate the BLEU4 scores. 

We use the official repository\footnote{\url{https://github.com/awslabs/mlm-scoring.git}} to calculate the Pseudo-Log-Likelihood scores of generated sentences.

\end{document}